\documentclass{article}
\PassOptionsToPackage{numbers}{natbib}
\usepackage[final]{neurips_2023}
\usepackage[utf8]{inputenc} 
\usepackage[T1]{fontenc}    
\usepackage{hyperref}       
\usepackage{url}            
\usepackage{booktabs}       
\usepackage{amsfonts}       
\usepackage{nicefrac}       
\usepackage{microtype}      
\usepackage{xcolor}         
\usepackage{graphicx}
\title{Comparative study of adapting pre-trained models for driving
behavior video captioning\thanks{Technical report based on the master thesis by the first author submitted 2024}}

\author{%
Sayak Mallick$^{2}$ \quad Philipp Geiger$^{1}$ \quad Augustin Kelava$^{2}$ \\
$^1$Bosch Center for Artificial Intelligence \quad $^2$University of Tübingen\\
\texttt{sayak.mallick@student.uni-tuebingen.de}\\
\texttt{Philipp.W.Geiger@de.bosch.com}\\
\texttt{augustin.kelava@uni-tuebingen.de}
}

\begin{document}

\maketitle

\begin{abstract}
This report examines and compares some of the many fine tuning and prompting methods existing, applying them within the domain of autonomous driving. The idea is to compare these methods by adapting a Large Language Model (LLM) on a video dataset. LLM's have become extremely good at achieving a good understanding of different forms of data and this study aims to induce a low dimensional understanding of driving situations into our primary test model SpaceTimeGPT. Experiments on BDD-X (Berkeley DeepDrive
eXplanation) dataset demonstrate good performance of the full fine tuning framework on some automatic metrics, and in some metrics, it even surpasses the baseline. We also try Low-Rank Adaptation (LoRA) and prompt engineering on VideoLLaVA model and discuss its limitations. 
\end{abstract}

\section{Introduction}
Driving video captioning is an important problem. Despite being effective, the lack of interpretability often limits our understanding of what goes on under the hood in the neural network approaches of these captioning models. An effective method of captioning and understanding can have several use cases in domain of Autonomous Driving. Firstly, this would allow text-based searching for interesting scenarios in a database. Secondly, the ability to express understanding in natural language would bridge the gap between an end user, who has limited knowledge of intelligent systems, and the machine’s behaviour. Thirdly, language can be looked at as a high-level representation for rule learning. There is a need for such representations and language is one such way to have short representations which nonetheless contain the things important for the decision making. Therefore, this could be an important step also towards better (representation) learning for driving. Fourth, this could make it easier to check to what extent a dataset of given scenarios covers a given operational design domain. We achieve this by using multiple encoder-decoder couples and cutting-edge LLM's for feature understanding in the videos. We also experiment with different adaptation methods and create a nice spectra of possible ways to achieve our objectives.

\section{Related work}

Notable related work in the domain includes ADAPT \cite{jin2023adaptactionawaredrivingcaption} which also uses the same dataset we use, BDD-X \cite{kim2018textual}, for their experimentation. 

ADAPT or Action-aware Driving Caption Transformer introduces a transformer architecture for captioning of driving videos. Just like our experiments, ADAPT addresses explainability and interpretability in this domain. They use a novel approach of jointly training driving caption task along with vehicular control prediction task by using a shared video representation. Since it is then expressed linguistically, it makes it easy to understand what is going on in the videos when compared to previous vision or LiDAR based approaches. The framework is then assessed on the BDD-X dataset. It results in state-of-the-art performance on automatic metrics and human evaluation. The paper includes a detailed ablation study analysing the impact of the different aspects of the design of this framework, different sampling rates and different control signal types. The authors also talk about the pipeline they developed to deploy this ADAPT framework in both simulator environments and the real world. This can facilitate real time conversion of raw driving video input into natural language. ADAPT is a direct motivation for us on this project. There were certain hurdles (eg. the docker setup limitations) which did not allow us to directly work with the ADAPT code and update it.

Wayve has released a research paper titled "GAIA-1: A Generative World Model for Autonomous Driving" \cite{hu2023gaia1generativeworldmodel} that explores the challenges of building autonomous driving systems, while focusing primarily on the problem of effectively predicting outcomes in response to the vehicle's actions in an evolving world. The paper introduces GAIA-1, a generative world model that leverages video, text, and action inputs to generate realistic driving scenarios while offering control over ego-vehicle behavior and scene characteristics.  Further, Wayve also released LINGO-2 which according to their website, is "the first closed-loop vision-language-action driving model (VLAM) tested on public roads". However, there is no publication available as of now. There is a previous paper on LingoQA \cite{marcu2024lingoqavideoquestionanswering}, which works towards the same direction.

\section{Setting and problem formulation}
\textbf{Setting:} In this paper, we consider video-to-text models having
\begin{itemize}
    \item the inputs $X$ which can be broken into visual signals $X_{pix}$ and further, optionally, textual input $X_{tex}$,
    \item a set of textual labels $Y$.
\end{itemize}

\textbf{Given:}
\begin{itemize}
    \item a general-purpose video captioning (video-to-text) model $M$ pre-trained on other large datasets (or individual components of them which can then be somehow connected, for eg. video encoder ${ENC}$ and text decoder ${DEC}$)
    \begin{itemize} 
        \item $\theta$ are all trainable parameters
        \item the model follows a probability distribution $p_\theta$ as its conditional density
        \item the density factorises as follows,

        \[p(Y\mid X_{pix},X_{tex}) = \prod_{i=1}^{L}p_\theta (Y^{[i]}\mid Y^{[1:i-1]},X_{pix},X_{tex}^{[1:i-1]}) \]
    \end{itemize}
    \item a dataset for the target task of driving video captioning $D$ (and potentially related side tasks) and $D$ $\sim$ $(X,Y)$ pairs.
      
\end{itemize}

\begin{figure}[htp]
    \centering
    \includegraphics[width=14cm]{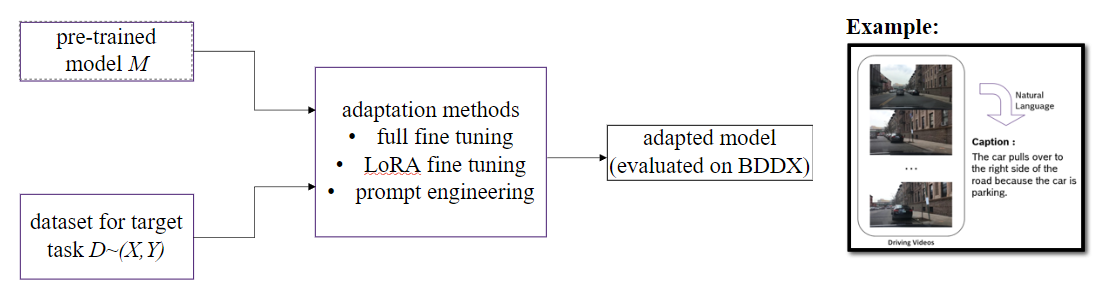}
    \caption{Summary of approach studied in this paper of how to get from dataset and pre-trained model to solution of the driving video captioning task (pictures taken from BDD / BDD-X).}
    \label{fig:approaches}
\end{figure}

\textbf{One of the goals:} to train a model that performs well on driving video behaviour captioning task. This can be achieved by maximizing the probability in the above equation. The objectives of the experiment change according to the approach taken.

The overview of this document is summarised in Figure \ref{fig:approaches}.

\section{Approaches}

The most important underlying mechanism for our experiments is the transformer architecture. Transformers have achieved state-of-the-art performance in various NLP tasks, including language translation, text summarization, and sentiment analysis, because of their ability to capture long-range dependencies and contextual information efficiently.

In the ViViT paper\cite{arnab2021vivitvideovisiontransformer}, using these kind of transformer models for video classification was studied. In order to handle the long sequences of tokens in video data, they propose many efficient variants of the model to factorise the spatial- and temporal-dimensions of the input.

The table in Figure 2 in the supplementary material sums up some key differences between classic text transformers and video transformers.

The approaches taken are discussed in two different sections: Pre-trained Models and Adaptation Methods.

\subsection{\textbf{Pre-trained Models:}}

A pre-trained model is a machine learning model that has already been trained on a large dataset for a particular task beforehand. They can be used as a starting point for other tasks by transferring their previous knowledge into the new problem at hand. This can be effectively achieved by fine-tuning them either completely or by freezing certain layers.

In our experimentation, we work with TimesFormer GPT-2, which is pretrained on the VaTeX (public dataset containing diverse video captioning and description instances)\cite{Wang_2019_ICCV}. The other model (VideoLLaVA) gets its pretraining datasets from multiple sources. The image pretraining is done on the LLAVA\cite{liu2023llava} dataset and the video pretraining is done on the VALLEY\cite{luo2023valley} dataset. The next sections summarises the different pretrained models that we use either off the shelf or with some sort of adaptation in our experimentation.

\subsubsection{TimeSformer GPT-2}
TimeSformer GPT-2 or TFGPT2 is an extension of the GPT-2 architecture integrating temporal information into its model, which is done by using the TimeSformer\cite{bertasius2021spacetimeattentionneedvideo} model. By incorporating time dependent structure $X_{pix}$ alongside textual data, This model enhances the understanding of sequential data. This makes it particularly useful for tasks involving time-series data or sequential information such as stock market prediction, video analysis, or natural language generation. The main idea is to develop upon the success of GPT-2 in generating coherent and contextually relevant text while also accounting for temporal dynamics, leading to improved performance in tasks which can benefit with understanding of time. In essence, this model is basically an encoder-decoder couple, where the encoder is "facebook/timesformer-base-finetuned-k600" from the Huggingface model repository and the decoder is "gpt2". In the TimeSformer paper\cite{bertasius2021spacetimeattentionneedvideo}, the authors have included a comparison between five space-time self-attention schemes: Space Attention, Joint Space-Time Attention, Divided Space-Time Attention, Sparse Local Global Attention and Axial Attention. Out of these, Divided Space-Time Attention achieves the best results. Compared to 3D CNN’s, the TimeSformer is a much improved model because CNN's have strong inductive biases like local connectivity and translation invariance. These characteristics are beneficial for small datasets but not in our use case. Local connectivity means that the neurons in a layer are connected only to a local region of input. Translation invariance means that a shift in the input image pixel values results in a proportionate shift in output pixel values. The TimeSformer has a large learning capacity in spite of its low inference cost. This facilitates increasing model capacity and means that the TimeSformer scales fairly well and is quicker to train compared to CNN’s\cite{olivastri2019endtoendvideocaptioning}. 

\subsubsection{TimeSformer BERT}
We also use the TimeSformer BERT model, which is basically the TimeSformer model in conjunction with the BERT language model\cite{devlin2019bertpretrainingdeepbidirectional}. It is slightly less powerful than the GPT-2 version, as we will see later in the experiments section.

\subsubsection{Video-LLaVA}
Video-LLaVA\cite{lin2023videollavalearningunitedvisual}, also known as Large-scale visual language is a model to better understand and process images and videos and is a powerful AI model designed to bridge the gap between how computers see and understand the world compared to humans. By converting video information into a format similar to text, Video-LLaVA can analyze videos like a large language model analyzes text. This allows it to deeply understand and consequently, grasp the content of a video as present in $X_{pix}$, including the actions and objects within it much more effectively than previous models. Additionally, a prompt $X_{tex}$ can be included. This allows for improved understanding which is key to applications like video summarization, generating captions, and answering questions based on video content. The training pipeline consists of two separate stages : image/video understanding and then instruction tuning.

\subsection{\textbf{Adaptation Methods}}

The two major adaptation methods we experiment with are the full fine tuning and the LoRA. We also try to engineer prompts for the bigger Video-LLaVA model.

\subsubsection{Full-fine tuning}

Classic full fine-tuning comprises of adjustment of the parameters $\theta$ of a pre-trained model to adapt it to a specific task by training it on the dataset $D$ for target task of driving video captioning. This process begins with a pre-trained model $M$ that has preciously been trained on a large and diverse dataset. Then, some layers may be frozen to encourage the model to remember key broad details from previous task. After this, the final layers (which presumably contain finer and task specific information) are replaced to fit the new task. This new network is then trained, using the new dataset $D$. This gives us a new set of weights and biases which are eventually used by the model to perform target task. An essential part of this is the use of an appropriate optimization algorithm (such as stochastic gradient descent) and the right loss function. Classic full fine tuning results in better performance on the new task than pretrained model\cite{dodge2020finetuningpretrainedlanguagemodels}. However, the process is often slow and requires significant compute power, especially for large models like the modern LLM's. There is also a high risk of overfitting in case the dataset is not large enough.

For classic fine tuning, the objective is
\[max_{\theta} \sum_ {D:(X,Y)} \sum_{i=1}^L log (p_{\theta} (Y^{[i]}\mid Y^{[1:i-1]},X_{pix},X_{tex}^{[1:i-1]}) \]

Let $\theta_{0}$ be all $\theta$ that are already present and $\Delta\theta$ be the new additional parameters to be trained. Usually, in any kind of fine tuning, the model is initialized with pretrained weights $\theta_{0}$ and updated to $\theta_{0}+\Delta\theta$.
In the case of full fine tuning, we do not learn a different set of parameters $\Delta\theta$. Therefore, $|\Delta\theta|$ here is $|\theta_{0}|$ as the initial parameters itself are further trained.

\subsubsection{Low-Rank Adaptation fine tuning}
LoRA fine-tuning\cite{hu2021loralowrankadaptationlarge} is a method that fine tunes more efficiently by introducing low-rank matrices into selected layers, while keeping most parameters fixed. This approach starts with a pre-trained model $M$. Most of the parameters stay fixed and hold the model's knowledge from the previous task. Then, we introduce low-rank matrices into some layers. This new architecture is then trained on the dataset $D$. These matrices capture the information that helps the model perform better on the new task. Like full fine-tuning, it is of prime importance to use the right optimization algorithm and loss function for the target task. Since we train only the newly added low rank matrices, while keeping the rest of the parameters fixed, LoRA is much more efficient computationally and also takes lesser time and memory. Further, there is also a lesser risk of over-fitting, since majority of the parameters are unchanged. In this way, LoRA allows us to quickly adapt pretrained models for new target tasks.

In the second part of the experiments, we implement LoRA on our target dataset $D$. This is a much more parameter-efficient approach, which allows the task specific parameter increment $\Delta\theta = \Delta\theta(\alpha)$ to be further encoded by a small number of parameters $\alpha$. The dimensionality of the parameter vector $\alpha$ is much smaller than the dimensionality of the original parameter vector $\theta_{0}$.
\[|\alpha| << |\theta_{0}|\]

Then, the revised objective is
\[max_{\alpha} \sum_ {D:(X,Y)} \sum_{i=1}^L log (p_{\theta_{0} + \Delta\theta(\alpha)} (Y^{[i]}\mid Y^{[1:i-1]},X_{pix},X_{tex}^{[1:i-1]}) \]

\subsubsection{Prompt Engineering and In-context learning}

In-context learning is an adaptation method where a pre-trained model learns how to perform better on a new target task by looking at examples of how its output should be. There are many different forms of in-context learning, like zero shot learning, one shot learning or many shot learning. In each of these, there is either zero, one or many examples included in the prompt, for the model to learn from. This, along with the query for the new task, leads the model to the right output. The approach allows the model to understand the context and generate responses based on the formats of the example/s provided. This method is frequently used as it hardly takes any further computational resources. The strong limitation of this method is that it is bound by the pre-existing knowledge of the model and the quality of examples in the prompt. 

Prompt engineering is an adaption method that improves performance on a task by customising the input prompts. In this type of adaptation, there is no change in the underlying parameters; instead, we leverage the pre-existing knowledge and logical understanding of the model. The prompt may be crafted to contain particular instructions or examples that nudges the LLM to outputting the desired behaviour based on task requirement. Prompt engineering requires no further computational resources as it only involves manipulation of the input text. However, as we see later on, the success of prompt engineering can be limited by the model's inherent capabilities and the quality of the crafted prompts.
\\
\\
On the TimeSformer-GPT2, we experiment with the classic fine tuning and the LoRA while on the larger Video-LLaVA model, we tried a couple of different prompt engineering techniques. The full fine tuning and LoRA for Video-LLaVA could not be completed due to time and computational restraints.

\section{Experiments}

We experiment with the BDD100K and the BDDX datasets for our tasks. They are both widely used datasets in the domain of autonomous driving.

\subsection{BDD100K}
BDD100K\cite{yu2020bdd100kdiversedrivingdataset}, or Berkeley Deep Drive 100K, is an important resource in the realm of computer vision and autonomous driving. It contains 100,000 videos, each about 40s long, and the format is 720p at 60 frames per second. The videos are collected and recorded all over the United States. This gives us a great mix of different circumstances such as city streets to countryside scenes. The dataset is labeled at a key-frame from every 10th second and have detailed annotations attached. This includes image tagging, bounding boxes and image segmentation details. There is also further information on road object detection. For our auxiliary task, we use this dataset to try and predict weather, surroundings and time of day columns as present in the dataset. We do this to gauge the level of understanding of the TimeSformer model. The results of this auxiliary task is presented in the supplementary section. 

\subsection{BDDX}
BDDX or Berkeley Deep Drive X \cite{kim2018textual}, is a unique dataset. It consists of video-label pairs and is one of the few datasets available that allows a model to learn captioning in the autonomous driving domain. BDDX has over 77 hours of driving videos spread across 7000 videos. These videos are tapes under a wide range of weather conditions such as rain, snow, sun. Just like the BDD100K, there is a plethora of videos for different situations, such as, overtakes, intersections, and simple speeding up. 
There is a pre-defined training/validation/testing split of the dataset. 
To process these videos in an easier manner, we preprocess dataset $D$ to contain video names directly and make each row a segment. In this form, dataset D has 5 total columns: a column containing video name, and the corresponding segment labels (start time, end time, caption, reasoning).

We can introduce BDDX as our primary dataset $D$ and break it down into its contents. The contents are:
-	the videos, X (the dataset itself contains video links which directly correspond to video names from a complete data pool of videos from BDD100K)
-	the corresponding labels, Y (consists of start time, end time, caption and reasoning for each segment of the video)

For our primary task, we try to predict captions that not only generate the action taken by the vehicle but also some sort of understanding as to why it takes it.

The four primary metrics we test on are the BLEU-4, CIDEr, METEOR and ROUGE-L. The CIDEr metric lies between 0 and 1000 while the rest are between 0 and 100.

Our first try is with the TFGPT2. In Table \ref{tab:table1}, we find the comparison between all the methods we experimented on.

\begin{table}[h] 
\centering 
\caption{Metric scores} 
\label{tab:table1} 
\resizebox{\textwidth}{!}{
\begin{tabular}{cccccccc} 
\toprule
Model used & Method of adaptation & Rank & Prompt Number & BLEU4 & CIDEr & METEOR & ROUGE-L \\ 
\midrule
ADAPT - Narration & - & - & - & 34.6 & 247.5 & 30.6 & 62.8\\
ADAPT - Reasoning & - & - & - & 11.4 & 102.6 & 15.2 & 32.0\\
\midrule
TF-GPT2 & Pretrained model & - & - & 0.02 & 3.9 & 7.9 & 12.4\\ 
TF-BERT & Pretrained model & - & - & 0.001 & 1.9 & 5.4 & 10.3\\ 
TF-GPT2 & Full fine-tuning & - & - & 10.0 & 105.9 & 34.9 & 44.3\\
TF-BERT & Full fine-tuning & - & - & 5.5 & 69.6 & 32.9 & 41.5\\
TF-GPT2 & LoRA fine tuning & 16 & - & 9.7 & 83.4 & 34.4 & 44.7\\
TF-GPT2 & LoRA fine tuning & 128 & - & 9.0 & 75.4 & 33.9 & 43.6\\
TF-GPT2 & LoRA fine tuning & 768 & - & 9.3 & 96.9 & 33.9 & 43.2\\
Video-LLaVA & Pretrained model & - & P1 & 0.02 & 1.6 & 19.3 & 17.9\\
Video-LLaVA & Prompt Engineering & - & P2 & 0.12 & 8.2 & 27.3 & 32.9\\
Video-LLaVA & Prompt Engineering & - & P3 & 0.09 & 6.7 & 26.5 & 31.2\\
\bottomrule
\end{tabular}
}
\end{table}

Initially, we compare the TF-GPT2 and TF-BERT pretrained models.Then comes the fully finetuned versions of the TFGPT2 and TF-BERT models. In the next part, we compare the different LoRA configurations for the TFGPT2 model. LoRA models can be modified by changing rank r and scaling factor alpha. We take a constant alpha and play around with the rank.

Then, we examine different prompt engineering techniques for the Video-LLaVA. The prompts that we experimented on and used are :
\\\textbf {Prompt number 1 P1:} "What action does the car take and why do you think it takes this action?"
\\\textbf {Prompt number 2 P2:} "What action does the car take and why? Please respond in one sentence in this format: 'The car takes (action) because (reasoning).' Include specific details about the traffic signals, road conditions, and any relevant environmental factors."
\\\textbf {Prompt number 3 P3:} "Describe the car's action and reasoning in one sentence in the following format: 'The car takes (action) because (reasoning).' Include details about the surrounding conditions and factors influencing the action."

The discussion of comparison between the methods is in the next section.

\section{Discussion and analysis}
\subsection {Discussion of methods}
\subsubsection{Pretrained models}
As expected, the pretrained models fall well short in the driving captioning and understanding task. This is because the data it is pretrained on is very different and having not looked at the new dataset, the model has no idea what to look for in the videos and structure of the video shot. In driving, video is always shot from the ego vehicle point of view. Here, we also see that the GPT2 version of TimeSformer is seemingly more expressive, because of the higher number of parameters present.

\subsubsection{Full fine tuning}

After first epoch of fine-tuning, the model seems to act as if the task is binary classification. With time and some more epochs, the model becomes more expressive and it becomes more like a multi-class classification task. It learns the most frequent outputs first, and the expressiveness and detail come only with the last few epochs of training.

\subsubsection{LoRA fine tuning}
Not all layers are trainable using LoRA. The first iterations using linear layers did not yield the desired outputs, as we saw repetitive outputs and unexpected sentence structures pop up. Finally, we settled on using linear, embedding and 1D and 2D convolution layers and this improved the outputs.

\subsubsection{Prompt engineering}
Prompt engineering in general seems to not perform too well here. We tried 3 different prompts ranging from simple to more complex. The performance definitely improves but it does not even come close to the outputs of the fine tuning.

\subsection {Discussion of results}
As expected, the full fine tuning outperforms the other experiments in all the metrics. It even outperforms some of the ADAPT metrics, which is appreciable because we achieve it with limited hardware and nonspecialist models made to understand the domain purely by adaptation. For the different LoRA, it seems that the model has issues with fitting as the different configurations seem to give similar results.

\section{Conclusion}
We see a glimpse of the power of the modern language models. Inspite of TimeSformer-GPT2 being much smaller and less complex than the ADAPT architecture, it performs comparably on many of the metrics. With further data, the capability of these models can be maximised. Also, more capable models like Video-LLaVA can be inspected and fine-tuned for driving video captioning and understanding. Although that does require significant computational resources and time, it should yield appreciable results.

We find that classical full fine-tuning, although computationally intensive yields the best results. The choice of adaptation method then becomes a conscious decision based on computational resources and time available to the user. Sometimes, simple prompt engineering can already lead to good results.

\appendix

\section{Appendix}

\subsection{Background on metrics}





\textbf{BLEU-4 : } 
Bilingual Evaluation Understudy\cite{papineni-etal-2002-bleu} is an inexpensive and language independent method of automatic machine translation evaluation that is used to evaluate quality of translated text by comparing it to one or more references. This is done by comparing n-grams in candidate translations to the references. BLEU-4 is the metric that focuses on 4-grams. It then takes the geometric mean and adds a brevity penalty to discourage shorter translation. Higher scores indicate better quality translations that closely match the reference texts in terms of word sequences.

\textbf{CIDEr : } 
Consensus-based Image Description Evaluation\cite{7299087} is a metric developed specifically for the evaluation of image captions generated by machine learning models. It measures the harmony, in a sense, between multiple reference captions and the generated caption. Unlike BLEU, which primarily focuses on n-gram overlaps, CIDEr considers both the similarity of individual words and their semantic relevance within the context of the image. In a way, it is n-gram similarity weighted with the TF-IDF score. Thus, this metric tends to correlate better with human judgments of caption quality, especially in scenarios where diversity and relevance are crucial.
 
\textbf{ROUGE-L : } 
Recall-Oriented Understudy for Gisting Evaluation - Longest Common Subsequence\cite{lin-2004-rouge} is a metric commonly used in natural language processing tasks, such as summarization or translation. It evaluates the quality of summaries or translations by computing the "longest common subsequence" between the candidate summary and the reference summaries. ROUGE-L emphasizes on recall by measuring exactly what part of the reference summary is captured in the candidate string. A higher scores means that there is a larger overlap between candidate and reference, and consequently results in a better quality of translation.

\textbf {METEOR : } 
Metric for Evaluation of Translation with Explicit Ordering\cite{banerjee-lavie-2005-meteor} is a metric that evaluates a translation by computing a score based on word-to-word matches between the translation and a reference. It considers both the lexical and the semantic similarities. It does so, by computing a harmonic mean of precision and recall, weighted by a measure of alignment between words in the candidate and reference. Additionally, METEOR takes into account stemming, synonyms and word orders while evaluating the final score, making it highly robust to variations in language expression. A higher METEOR scores indicate better quality translations that capture both surface-level and semantic aspects of the reference text.

\bibliographystyle{unsrtnat}
\bibliography{bibliography.bib}

\end{document}